\documentclass[12pt]{article}
\usepackage{amsmath}
\usepackage{graphicx}
\usepackage{enumerate}
\usepackage{comment}
\usepackage{placeins} 
\usepackage{natbib}
\usepackage{url} 
\usepackage{titlesec}
\titlelabel{\thetitle.\quad}
\newcommand{\blind}{1}

\RequirePackage[OT1]{fontenc} \RequirePackage{amsthm,amsmath}
\usepackage{graphics,epsfig,epsf,psfrag}
\usepackage{url}
\usepackage{amsmath,amsfonts,amssymb,amsthm}
\usepackage{bbm, bm}
\usepackage{graphicx,lscape,rotate}
\usepackage{movie15}  
\usepackage{dashrule} 
\usepackage{epstopdf} 
\usepackage{enumerate, framed, afterpage}
\usepackage{colortbl}
\usepackage{color, appendix}

\RequirePackage[colorlinks,citecolor=blue,urlcolor=blue]{hyperref}
\usepackage{pstricks,pst-node,pst-tree}  
\usepackage{algorithmic} 
\usepackage[linesnumbered,lined,commentsnumbered]{algorithm2e}
\usepackage{dsfont,setspace,subcaption}
\usepackage{float}
\newcommand{\algrule}[1][1.pt]{\par\vskip.3\baselineskip\hrule height #1\par\vskip.3\baselineskip}

\usepackage{graphics, hyperref}  

\newtheorem{theorem}{Theorem}
\newtheorem{corollary}{Corollary}
\newtheorem{lemma}{Lemma}
\newtheorem*{remark*}{Remark} 
\theoremstyle{definition}
\setcitestyle{citesep={;}}

\newcommand{\sgn}{\mbox{sgn}}

\newcount\colveccount
\newcommand*\colvec[1]{
        \global\colveccount#1
        \begin{pmatrix}
        \colvecnext
}
\def\colvecnext#1{
        #1
        \global\advance\colveccount-1
        \ifnum\colveccount>0
                \\
                \expandafter\colvecnext
        \else
                \end{pmatrix}
        \fi
}

\makeatletter
\newcommand{\Spvek}[2][r]{%
  \gdef\@VORNE{1}
  \left(\hskip-\arraycolsep%
    \begin{array}{#1}\vekSp@lten{#2}\end{array}%
  \hskip-\arraycolsep\right)}

\def\vekSp@lten#1{\xvekSp@lten#1;vekL@stLine;}
\def\vekL@stLine{vekL@stLine}
\def\xvekSp@lten#1;{\def\temp{#1}%
  \ifx\temp\vekL@stLine
  \else
    \ifnum\@VORNE=1\gdef\@VORNE{0}
    \else\@arraycr\fi%
    #1%
    \expandafter\xvekSp@lten
  \fi}
\makeatother

\begin{document}

\def\spacingset#1{\renewcommand{\baselinestretch}%
{#1}\small\normalsize} \spacingset{1}


\if1\blind
{ 
  \title{\bf  Conservative Decisions with Risk Scores}
  \author{ \textbf{Yishu Wei} \\
    Reddit Inc., San Francisco, CA 94102, USA \vspace{0.1in} \\ 
    \textbf{Wen-Yee Lee} \\
    Department of Chemistry \& Biochemistry \\ 
    University of Texas, El Paso, TX, 79968, USA \vspace{0.1in} \\
    \textbf{George Ekow Quaye}  \\
    Biostatistics and Epidemiology Core \\
    Health Services and Outcomes Research \\
Children’s Mercy Research Institute, MO 64108, USA \vspace{0.1in}  \\
and  \textbf{Xiaogang Su} \\
    Department of Mathematical Sciences \\
    University of Texas, El Paso, TX, 79968, USA
    \hspace{.4cm}
    }
    
      \maketitle
} \fi

\if0\blind
{
  \bigskip
  \bigskip
  \bigskip
  \begin{center}
    {\LARGE\bf Conservative Decisions with Risk Scores in Prostate Cancer Diagnosis}
\end{center}
  \medskip
} \fi

\begin{abstract}
In binary classification applications, conservative decision-making that allows for abstention can be advantageous. To this end, we introduce a novel approach that determines the optimal cutoff interval for risk scores, which can be directly available or derived from fitted models. Within this interval, the algorithm refrains from making decisions, while outside the interval, classification accuracy is maximized. Our approach is inspired by support vector machines (SVM), but differs in that it minimizes the classification margin rather than maximizing it. We provide the theoretical optimal solution to this problem, which holds important practical implications. Our proposed method not only supports conservative decision-making but also inherently results in a risk-coverage curve.  Together with the area under the curve (AUC), this curve can serve as a comprehensive performance metric for evaluating and comparing classifiers, akin to the receiver operating characteristic (ROC) curve. To investigate and illustrate our approach, we conduct both simulation studies and a real-world case study in the context of diagnosing prostate cancer.
\end{abstract}

\noindent%


\section{Introduction}
\label{sec-Introduction}

In safety-critical applications, the stakes are high, and a binary classification can incur severe consequences if an erroneous decision is made. This applies to both false positives and false negatives, both of which can lead to significant harm. To mitigate these risks, conservative decision-making that allows for abstention can be beneficial. By abstaining from making a decision in cases of high uncertainty, the potential losses can be reduced and the overall safety of the system can be improved.

Take prostate cancer diagnosis for example. Prostate cancer is a widespread cancer that affects men globally. Early detection is crucial for effective treatment and improving survival rates. However, the diagnosis of prostate cancer can be challenging. One common risk score for prostate cancer screening is the Prostate-specific antigen (PSA) test \cite[e.g.,][]{fenton2018prostate}. PSA is a protein produced by the prostate gland, and a PSA test measures the level of this protein in the bloodstream. While elevated PSA levels can be an indication of prostate cancer, it is important to note that non-cancerous conditions can also cause an increase in PSA levels. Consequently, PSA testing has limitations in terms of sensitivity and specificity, and false positive and false negative errors are common \citep{van2012epigenetic}. 

Conservative decision-making approaches prioritize minimizing incorrect diagnoses while enabling effective decision-making for the majority of cases. In statistical and machine learning contexts, such approaches are commonly referred to as classification with a reject option. Early work in this area dates back to \citet{chow1970optimum}, who proposed a cost-based framework that incorporates an abstention or `reject' option with a predetermined cost and showed that the optimal solution is a modified Bayes classifier. Various learning methods can be used to learn the optimal strategy. One popular approach is support vector machines (SVM) with a reject option \citep{fumera2002support,grandvalet2008support,wegkamp2011support}. However, the cost-based model requires explicit definition of the abstention cost, which can be challenging to determine in practical applications. 

In order to address the problem, \citet{pietraszek2005optimizing} proposed an alternative framework that avoids explicitly considering the reject cost. This framework involves evaluating two antagonistic quantities: the selective risk, which is the risk associated with decision-making, and the coverage, which is the probability that a decision can be made. These considerations lead to two approaches: bounded-improvement and bounded-abstention. The bounded-improvement model seeks to maximize coverage while maintaining a guaranteed selective risk, while the bounded-abstention model aims to minimize selective risk while maintaining a guaranteed coverage. In a related vein, previous works have explored different aspects of this framework. \citet{el2010foundations} investigated the noise-free setting, \citet{geifman2017selective} proposed an algorithm for the context of deep learning, and \citet{lei2014classification} studied a bounded-abstention model with separate control over true-positive and true-negative coverage probabilities. More recently, \citet{franc2023optimal} demonstrated that optimal solutions typically resemble the modified Bayes classifier in the cost-based framework. 

The determination of the optimal cutoff point for bisecting a risk score, according to various criteria, is discussed in \cite{lopez2014optimalcutpoints} and  references therein. This cutoff point is critical for converting a continuous prediction or risk score into a binary decision without abstention. \citet{zhang2020bootstrap} propose an infinitesimal jackknife approach to compute the standard errors of the estimated optimal cutoff point. The confidence interval derived in \citet{zhang2020bootstrap}  can be used in decision-making processes with abstention. Specifically, observations with risk scores falling within the confidence interval can be rejected or withheld, offering a practical way to handle uncertainty when converting a continuous prediction or risk score into a binary decision.

This article presents a novel approach to conservative decision-making on risk scores, building on the frameworks of \citet{pietraszek2005optimizing} and \citet{franc2023optimal}. Our proposed approach addresses scenarios where risk scores may be directly available or learned from data. We introduce an optimal cutoff interval that can be naturally formulated as a convex programming problem, inspired by the support vector machine (SVM) approach. But our method differs from SVM in the context of conservative decision-making, where the classification margin is minimized rather than maximized. We obtain the theoretical solution for the optimal cutoff interval, which is related to the Bayes classifier as discussed in \citet{franc2023optimal}. However, our approach differs in that the selective function and coverage function are both determined by the same cutoff interval. As a result, our derivation incorporates distinct conditions and approaches. Another noteworthy result of our method is that the optimal cutoff interval enhances the practical usefulness of the risk-coverage (RC) curve and the area under the curve (AUC). By integrating our method, these metrics can now be routinely used as comprehensive performance measures to assess and compare classifiers, akin to the receiver operating characteristic (ROC) curve. 

The remainder of the article is organized in the following manner. Section \ref{sec-method} presents the proposed method for seeking the optimal cutoff interval. The theoretical solution is obtained in Section \ref{sec-theory}. Section \ref{sec-simulation} investigates the performance of our method through simulation studies. In Section \ref{sec-example}, an illustration of our method in prostate cancer diagnosis is provided. Finally, Section \ref{sec-discussion} concludes the article with a brief discussion.

\section{Optimal Cutoff Interval}
\label{sec-method}
The available data $\mathcal{D}$ consist of $\{(y_i, r_i) \in \mathcal{Y} \times \mathbb{R}: i = 1, \ldots, n\}$, where the response $y_i$ is binary with support $\mathcal{Y} = \{ \pm 1\}$ and  $r_i$ is the risk or uncertainty score. It is worth noting that the risk score $r_i$ itself could be either directly available or learned from data. Without loss of generality, we assume that positive cases tend to be associated with larger risk scores than negative cases. Our goal is to find an interval $(c-d, c+d)$ with center $c$ and half width $d \geq 0$ such that observations with risk scores falling outside the interval are all or mostly classified correctly whereas decisions on observations within the interval are withheld. 

We start with the hard-margin scenario where it is required that every observation outside the interval must be correctly classified. In this case, clearly we want the interval to be the narrowest. Thus the problem may be formulated as 
\begin{equation}
\label{opt0-hardmargin}
\min_{c, d}~ d, \mbox{~~~~~~~~s.t.~} d \geq 0 \mbox{~and~} - y_i ( r_i -c ) - d \leq 0 ~~ \forall i,
\end{equation}
where the constraint $- y_i ( r_i -c ) - d$ or, equivalently, $y_i ( r_i -c ) + d \geq 0$ is a unified way of expressing the two conditions that $r_i \geq c -d $ if $y_i = +1$ and $r_i \leq c+ d$ if $y_i = -1.$   

\begin{figure}[h!]
\centering
  \includegraphics[scale=0.62, angle=0]{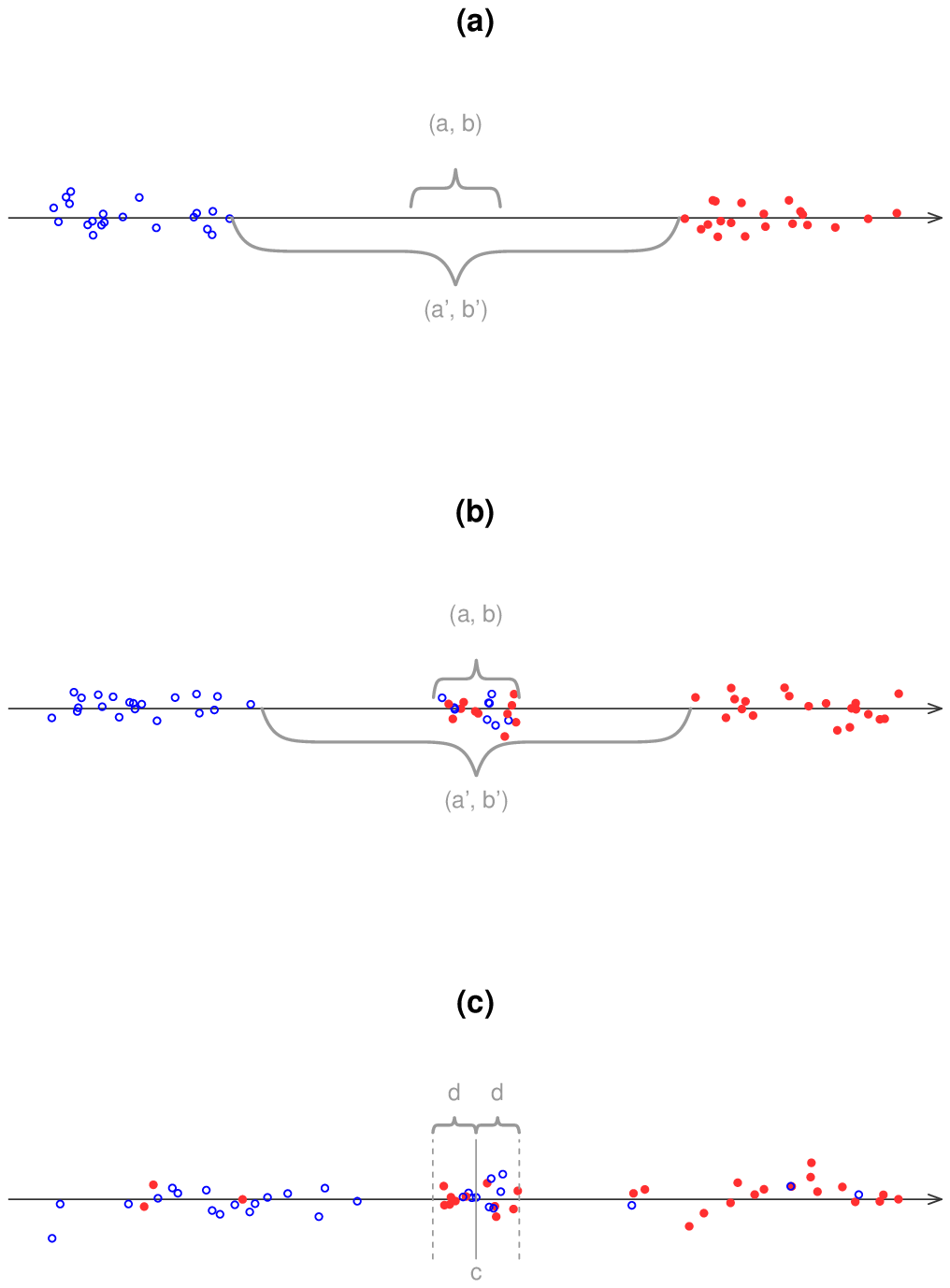}
  \caption{Illustration of SVM (a) and conservative decision with hard (b) and soft (c) margins based on risk scores. Positive cases are denoted as solid dots while negative cases are denoted as circles. \label{fig01}}
\end{figure}

Instead of maximizing the margin as in SVM, the margin $d$ is minimized in our formulation. This is because we are classifying with a band or interval and, though reluctantly, allow for indecision with observations that are hard to classify. Figure \ref{fig01} provides a further illustration. Panel (a) illustrates the binary decisions in SVM, where $[a', b']$ with a larger margin is preferable to $[a, b]$ in achieving maximal separation of positive and negative cases. Panel (b) presents our scenario with ternary decisions. Clearly, interval $[a, b]$ with a shorter width is better than $[a', b']$, since we want to keep the number of cases with indecision small. This idea aligns with the bounded-improvement framework \citep{pietraszek2005optimizing, franc2023optimal}, where $1-\Pr\{r \in (a, b)\}$ is referred to as coverage -- the probability of being able to make a decision.

Solving (\ref{opt0-hardmargin}) would estimate $c-d$ as the lowest risk score with a positive response and $c+d$ as the highest risk score with a negative response. This approach would be sensitive to outliers that are misclassified, which is undesirable. See Figure \ref{fig01}(c) for an illustration. A more practical approach is to allow for some misclassifications outside the interval. With the Tikhonov approach, we penalize on those observations. This leads to the following formulation
\begin{equation}
\label{opt-softmargin}
\min_{c, d}~ d  + \gamma \cdot \sum_{i=1}^n \max\{0, ~ - y_i ( r_i -c ) - d \},
\end{equation} 
where the tuning parameter $\gamma \geq 0$ represents the rate of penalty or cost for each misclassification outside the interval.

In the above Tikhonov formulation, a penalty is applied only when $- y_i ( r_i -c ) - d >0$, the scenario when a mislclassification occurs outside the interval. Although (\ref{opt-softmargin}) has a simple form, the max function is not differentiable. To gain differentiability, we introduce slack variables $ \xi_i = \max\{0, ~ - y_i ( r_i -c ) - d \}$ for $i=1, \ldots, n$ and rewrite the problem as 
\begin{eqnarray}
& \min_{c,d, \bm{\xi}} ~~ d + \gamma \cdot \sum_{i=1}^n \xi_i \nonumber \\
\mbox{s.t.~} & - d \leq 0, ~ - \xi_i \leq 0, \mbox{~and~} -y_i(r_i-c) -d - \xi_i \leq 0, \forall i.  \label{opt-primal}
\end{eqnarray}
This is a convex programming \citep{boyd2004convex} problem in its standard form.

The corresponding Lagrangian is 
\begin{equation}
\label{Lagrangian}
\mathcal{L}(c, d, \bm{\xi}_i; \alpha, \bm{\lambda}, \bm{\mu}) = d + \gamma \cdot \sum_{i=1}^n \xi_i - \alpha \, d - \sum_{i=1}^n \lambda_i \xi_i - \sum_{i=1}^n \mu_i A_i,
\end{equation}
where $\alpha, \lambda_i, \mu_i \geq 0$ are the Lagrangian multipliers or dual variables and the term $$A_i = y_i (r_i-c) + d + \xi_i$$ is introduced to simplify notations. 

We next look into the KKT conditions. First, the stationarity condition gives
\begin{equation}
\left\{ 
\arraycolsep=1.4pt\def\arraystretch{1.2}
\begin{array}{l}
 \partial \mathcal{L}/\partial c = \sum_{i=1}^n \mu_i y_i = 0 \\
 \partial \mathcal{L}/\partial d = 1- \alpha - \sum_{i=1}^n \mu_i = 0  ~~ \Longleftrightarrow ~~ \sum_{i=1}^n \mu_i = 1- \alpha \\
 \partial \mathcal{L}/ \partial \xi_i = \gamma - \lambda_i - \mu_i = 0 
~~ \Longleftrightarrow ~~   \lambda_i + \mu_i = \gamma, \, \forall i 
\end{array}
\right. \label{stationarity}
\end{equation}
The primal feasibility conditions are $d \geq 0$, $\xi_i \geq 0$, and $y_i(r_i-c) + d + \xi_i \geq 0 ~ \forall i$. The dual feasibility conditions are $\alpha \geq 0$, $\lambda_i \geq 0$, and $\mu_i \geq 0.$ 
Finally, the complementary slackness conditions implies that 
\begin{equation}
\label{complementary-slackness}
\alpha d =0, ~ \lambda_i \xi_i = 0, \mbox{~and~} \mu_i A_i =0. 
\end{equation}

A case analysis is induced by complementary slackness conditions. First of all, if $\alpha >0$, then $d =0.$ The optimal cutoff interval problem would reduce to the optimal cut off point \citep{lopez2014optimalcutpoints} that minimizes the misclassification rate. Thus we focus primarily on scenarios with $d >0$, which implies $\alpha =0.$  One condition in (\ref{stationarity}) $\sum_{i=1}^n \mu_i = 1- \alpha =1$ implies that $\mu_i$'s are weights.  

One interesting scenario is when $0 < \lambda_i < \gamma$ or $0 < \mu_i < \gamma.$ In this case, we must have $\xi_i =0$ and $A_i =0.$ This occurs to positive observations with $r_i = c-d$ and negative observations with $r_i = c+d$. Namely, these observations fall right on the two boundaries. They are called support vectors in SVM. Another interesting scenario is when $A_i >0$. This implies $\mu_i =0$ and hence $\lambda_i  = \gamma$, which in turn implies that $\xi_i = 0.$ Put together, we must have $y_i (r_i -c) + d >0$. These are those correctly classified observations outside the interval, each with $\mu_i =0.$ Thus $\mu_i$'s are sparse weights. Yet another interesting scenario is when $\xi_i >0.$ This implies $\lambda_i =0$ and $\mu_i = \gamma$. Hence $A_i =0$, i.e., $y_i(r_i-c)+d + \xi_i =0.$ This scenario corresponds to misclassified cases outside the interval.

Bringing the KKT conditions into the Lagrangain and rewriting leads to the following Wolfe's dual form \citep{wolfe1961duality}:
\begin{equation}
\label{opt-dual}
\arraycolsep=1.4pt\def\arraystretch{1.5}
\begin{array}{ll}
& \min_{\bm{\mu}} ~ \sum_{i=1}^n y_i r_i \mu_i \\
\mbox{s.t.~} & 0 \leq \mu_i \leq \gamma, ~~ \sum_{i=1}^n \mu_i = 1, \mbox{~~and~} \sum_{i=1}^n y_i \mu_i =0.
\end{array}
\end{equation}
The dual problem presented above is a convex programming problem. It can be easily verified that strong duality holds for both (\ref{opt-primal}) and (\ref{opt-dual}). Thus we solve the dual problem first for $\mu_i$s. 
For a detailed description of the estimation algorithm, see Algorithm 1 in Section 2 of the Supplementary Materials. 

To determine the optimal tuning parameter $\gamma$, two graphical plots can be informative. First, we plot the classification accuracy and coverage, each as a function of $\gamma$. An appropriate choice of $\gamma$ should correspond to a scenario where both accuracy and coverage are high. In the second plot, the misclassification error rate is plotted as a function of coverage, commonly known as the risk-coverage (RC) curve in the literature \citep{corbiere2019addressing, franc2023optimal}. To identify a suitable value for $\gamma$, one can look for kinks in the plot where the misclassification error remains constant as the coverage increases above a certain threshold. The area under the risk-coverage  curve (AUC) can serve as a comprehensive performance metric, capturing the overall effectiveness of the risk score in accurately classifying responses. 

Our proposed method enhances the practical usefulness of the risk-coverage curve. We advocate for its routine use in the assessing and comparing classifiers, akin to the receiver operating characteristic (ROC) curve. It is noteworthy that the two AUC values carry different interpretations. The area under the ROC curve, also known as the C-statistic or concordance, serves as a measure of discrimination or separation between patients and non-patients. Specifically, it reflects the probability that a randomly selected patient is ranked higher than a randomly chosen non-patient. In contrast, the area under the RC curve is more straightforward to interpret. It represents the average optimal risk as coverage varies from 0 to 1, with optimality supported by our method.

\section{Theoretical Solution}
\label{sec-theory}

We study the population version of the problem and derive the corresponding optimal solution $(c^\star, d^\star),$ which has interesting practical implications. To set up, note that problem (\ref{opt-primal}) is the penalized form of the constrained problem 
\begin{equation}
\label{opt-constaint-primal}
\min_{c, d}~~ \frac{1}{n} \sum_{i=1}^n \max\{0, - y_i(r_i-c) -d \}, ~~\mbox{s.t.}~  d = t,
\end{equation}
with $t \geq 0$ being a tuning parameter, which can be viewed as a surrogate of the following problem 
\begin{equation}
\label{opt-misclassification}
 \min_{c, d}~~ \frac{1}{n}\sum_{i=1}^n I\left\{ y_i(r_i-c)  \leq - d \right\}, ~~\mbox{s.t.}~  d = t. 
\end{equation}
The indicator function in (\ref{opt-misclassification}) is replaced with the hinge function in (\ref{opt-constaint-primal}) as a convex relaxation. 
\citet{bartlett2006convexity} provided a general study of the surrogate function. In (\ref{opt-misclassification}), the misclassification error is  minimized, subject to the constraint on $d$. This is in general a bounded-improvement model in the sense of \citet{franc2023optimal}. The bound on $d$ is essentially to control the coverage probability, which refers to the probability when an affirmatory decision can be made.

These considerations motivate us to consider the following population problem 
\begin{equation}
\label{opt-population} 
\min_{c, d}~~ \Pr \left\{ y (r-c) \leq -d  \right\}, ~~\mbox{s.t.}~
\Pr( |r-c| > d ) = \theta,
\end{equation}
where $0 \leq \theta \leq 1$ is a fixed probability of coverage. We assume that $\theta$ is large, e.g., 80\%, 90\%, 95\%, 99\%, since we would prefer to have the capacity to make decisions for most of the time. Let $\widehat{y}_{c, d}$ be the decision value on the basis of $c$ and $d$, i.e., 
$$ \widehat{y}_{c, d} =  \widehat{y}_{c, d}(r) = \left\{
\begin{array}{ll}
-1 & \mbox{if~~} r < c - d, \\
0 & \mbox{if~~} |r-c| \leq d, \\
+1 & \mbox{if~~} r > c + d,
\end{array}
\right.
$$
where $\widehat{y}_{c, d} =0 $ corresponds to the `abstention' or `reject' scenario. The loss function is 
$$ l(y, \widehat{y}) = I\left( y \widehat{y} <0 \right). 
$$
The resultant risk function is 
$$ R(\widehat{y}_{c,d}) = E_{y, r} l(y, \widehat{y}_{c,d}) = \Pr \left\{ y (r-c) \leq -d  \right\},$$
which amounts to the misclassification error rate. In other words, (\ref{opt-population}) minimizes the misclassification error while keeping the coverage probability fixed at some $\theta$ for $0 \leq \theta \leq 1.$ 

\begin{theorem}
\label{theorem-1}
Assume that $\pi(r) = \Pr(y = +1 | r)$ is a monotone increasing function of the risk score $r$. Define $c^*$ as the value of $r$ such that  $ \pi(c^\star) = 0.5$ and $d^\star$ as the threshold such that $\Pr( |r-c^\star| > d^\star ) = \theta.$ Additionally, we assume that $\pi(r)$ satisfies the following symmetry condition around $c^\star$:
\begin{equation}
\label{symmetry}
\pi(c^* - a) = 1- \pi(c^* + a), ~~ \forall \, 0 \leq a \leq d^\star.
\end{equation} 
Then $(c^\star, d^\star)$ is an optimal solution of (\ref{opt-population}).
\end{theorem}

The proof is deferred to the Supplementary Materials. The problem  (\ref{opt-population}) is generally a bounded-abstention model as described in \citet{franc2023optimal} and Theorem \ref{theorem-1} resembles their optimal solutions. However, our decision function and coverage function are both determined by $c$ and $d$. As a result, their arguments are not applicable. A different approach is taken in our proof; see Section 1 in the Supplementary Materials. 

Recall that the Bayes classifier 
$$\widehat{y}_B = \sgn \left[ \Pr(y = +1 | r) - 0.5 \right]$$
is the optimal classifier when minimizing the risk of 0-1 loss without abstention. Theorem \ref{theorem-1} demonstrates that, under reasonable conditions, the optimal cutoff point $c^\star$ for decision-making with abstention remains the same as that of the Bayes classifier. The optimal half-width $d^\star$ can then be determined accordingly. 

The condition of symmetry around $c^\star$ for $\pi(r)$ is essential in guaranteeing that the classifiers being analyzed conform to the abstention format of $(c-d, c+d)$. At first glance, this requirement may appear overly strict. The subsequent lemma shows that $\pi(r)$ can be easily adjusted to satisfy this symmetry property through transformation. Besides, its property of monotonicity is well preserved after transformation.   
\begin{lemma}
\label{lemma}
Assuming that $\pi(r)$ is monotone increasing and continuous, there must exist another  monotone increasing continuous function $\pi'(r)$ that is bounded within $(0,1)$ and satisfies:
\begin{enumerate}[(i).] 
\item $\pi'(c^\star) = 0.5  \mbox{~iff~} \pi'(c^\star)= 0.5$;
\item $\pi'(r)$ satisfies the symmetry property (\ref{symmetry}) around $c^\star$;
\item $ \pi(r) \geq \pi(r') \mbox{~iff~} \pi'(r) \geq \pi'(r')$. 
\end{enumerate}
\end{lemma}
Hence, we can conclude that the symmetry condition is not a significant constraint on $\pi(r)$. A risk score $r$ must be monotonic with respect to the underlying risk probability $\pi(r)$ in order to qualify as a valid risk score. This requirement allows for a wide range of possibilities when defining a risk score.

Next, we consider the scenario when $r$ is developed from data $\{(y_i, \bm{x}_i): i=1, \ldots, n\}$ which consist of $n$ IID copies of $(y, \bm{x}) \in \mathcal{Y} \times \mathcal{X}$ with $\mathcal{X} \subseteq \mathbb{R}^p.$ To obtain a risk score in this classification problem, one can regress $y$ on $\bm{x}$. Various classifiers, such as logistic regression, are available for this purpose. Typically, the risk score $r$ is a monotonic function of $\eta(\bm{x}) = \Pr(y=1|\bm{x})$, and vice versa. 
\begin{corollary}
\label{corollary}
Assume that $\eta(\bm{x}) = \Pr(y = +1 |\bm{x}) = \pi(r)$ is a monotone increasing function of the risk score $r$. We define $c^*$ as the threshold value such that $r \leq c^\star $ if and only if $\eta(\bm{x}) \leq 0.5$, and $d^\star$ as the value such that $\Pr( |r-c^\star| > d^\star ) = \theta$. Further assuming that $\pi(r)$ satisfies the same symmetry property around $c^\star$ as described in Theorem \ref{theorem-1}, we can conclude that the pair $(c^\star, d^\star)$ represents an optimal solution to (\ref{opt-population}).
\end{corollary}

Corollary \ref{corollary} expands on the results of Theorem \ref{theorem-1} to scenarios where the risk score has not yet been determined. The proof follows directly from Theorem \ref{theorem-1} and is therefore omitted. Corollary \ref{corollary} has interesting practical implications. In practical implementation of the proposed decision procedure, it is intuitive to iterate between estimating the risk score $\eta(\bm{x}) = \pi(r)$ and estimation of the cutoff interval $\{c, d\}.$ Nevertheless, Corollary \ref{corollary} shows that the iteration is unnecessary, since the optimal choices of $\{\pi(r), c, d\}$ are fully determined. On the other hand, the conclusions in Corollary \ref{corollary} rely on certain assumptions. Firstly, we assume independent and identically distributed (IID) observations. However, when decision-making challenges arise from a contamination distribution that substantially differs from the main data distribution, multiple iterations can improve the estimation of the risk score and cutoff interval. We will investigate this through a simulation study. Secondly, the symmetry assumption for $\pi(r)$ is not naturally met by risk scores in practical scenarios, although it can be attained through transformation. Therefore, it is essential to estimate both $c$ and $d$ based on the original risk scores.

\section{Simulation Studies}
\label{sec-simulation}
Our simulation experiment comprises three studies. First, we evaluate the proposed method using two scenarios illustrated in Figure \ref{fig01}: an ideal separation case and an ideal separation case with added noise, serving as a sanity check. Second, we apply the method to risk scores derived from logistic regression, examining the trade-off between coverage and accuracy under various settings. Finally, we investigate a two-step SVM approach, showing that fitting on non-rejected instances enhances the estimation of the classification boundary in noisy conditions. The performance measures used are: $Coverage$ (proportion of non-rejected samples), $acc_{nonreject}$ (accuracy on non-rejected points), and $acc_{all}$ (accuracy on all samples).

\subsection{Study I: Ideal Separation without and with Noise}
\label{sec-study1}

We first examine the ideal separation scenario in Figure \ref{fig01}(b) as a sanity check for our method. To generate data, we simulate  
positive cases from uniform $[0.4, 1]$ and negative cases from uniform$[0, 0.5]$ and then remove observations falling in intervals $[0.3, 0.4]$ and $[0.5, 0.6]$. Thus there is an overlapping area where positive and negative instances mix along the risk score axis. Outside this area, there is a gap without any data points, followed by complete separability. 

\begin{figure}[H]
     \centering
     \begin{subfigure}[b]{1.0\textwidth}
         \centering
         \caption{}
         \includegraphics[width=1.\linewidth, height=0.3\textheight]{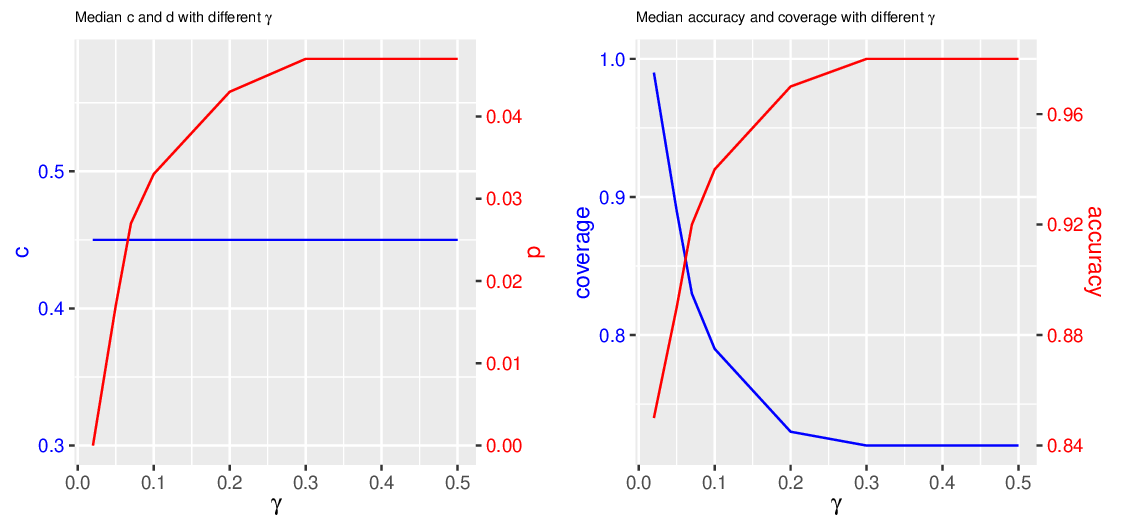}
     \end{subfigure}
     \begin{subfigure}[b]{1.0\textwidth}
         \centering
          \caption{}
         \includegraphics[width=1.0\linewidth, height=0.3\textheight]{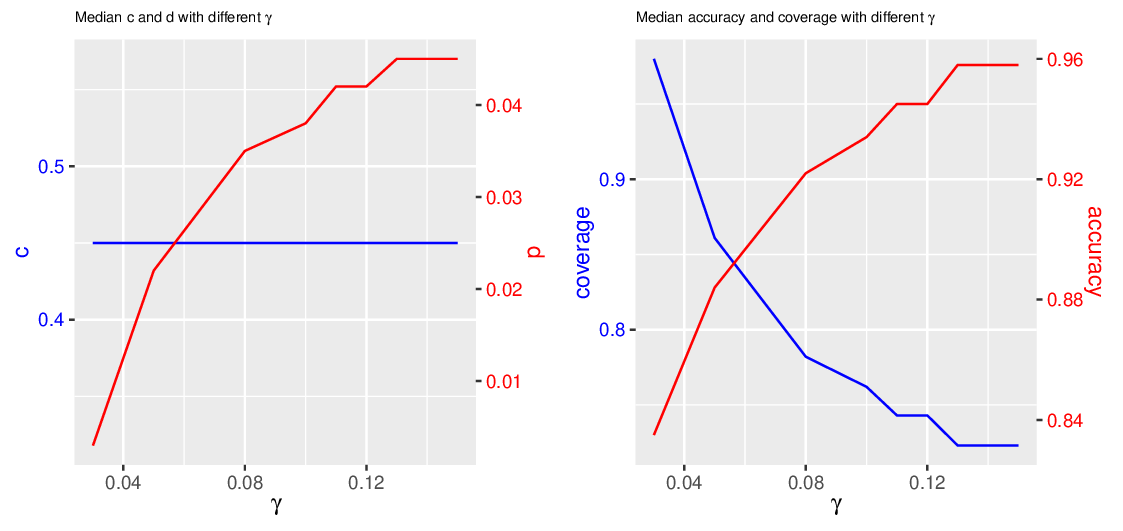}
     \end{subfigure}
        \caption{Ideal separation case without (a) and with (b) noise, examined in Section \ref{sec-study1}. The results are based on the median values of $c$ and $d$ estimates, as well as the coverage and accuracy, obtained from 200 simulation runs.}
        \label{fig02}
\end{figure}

Figure \ref{fig02}(a) summarizes the results from 200 simulation runs with data sets of size $n=200$. In the left panel, the (median) estimated center $c$ and half interval length $d$ are plotted as a function of the penalization factor $\gamma$, over the simulation runs. The plot demonstrates that the estimated $c$ remains consistently close to the ground truth value of 0.45, while $d$ gradually increases to the true value of 0.05 and remains stable as $\gamma$ increases. The right panel illustrates that the coverage decreases as $\gamma$ increases, while the accuracy reaches 100\%. At $\gamma=0.3$, the estimated overlapping area is $[0.403, 0.497]$, which closely approximates the ground truth value of $[0.4, 0.5]$.

The results demonstrate that our proposed method accurately identifies the true overlapping interval in the ideal separation case. By adjusting $\gamma$ in (\ref{opt-primal}), our method successfully identifies the overlapping area $[0.4, 0.5]$ while excluding the gap. The estimated values of $c$ and $d$ closely align with their true values of $c = 0.45$ and $d = 0.05$. Notably, the presence of gaps affects both the value of $d$ and the coverage as $\gamma$ exceeds 0.3, highlighting the practical implications for selecting the tuning parameter $\gamma$.

Next, we examine the ideal separation case with added noise, as illustrated in Figure \ref{fig01}(c). The simulation setup is the same as the ideal separation case, except that 3\% of positive and negative instances are flipped. The results are presented in Figure \ref{fig02}(b). The plot shows that the estimated center $c$ consistently remains close to the true value of 0.45. As the penalty parameter $\gamma$ increases, both the interval half-length $d$ and coverage gradually approach their true values, exhibiting an `elbow' shape in the curve. This pattern highlights the effectiveness of our proposed method. Similar to the previous scenario, the presence of an `elbow' in the coverage and $d$ curves helps determine an appropriate value for $\gamma$. Additionally, unreported simulation results indicate that our SVM-based approach exhibits robustness to the distributions of risk scores. 

\subsection{Study II: Risk Score from Logistic Regression}
\label{sec-study2}

In the next study, we investigate risk scores generated from logistic regression models and examine the tradeoff between accuracy and coverage under different configurations of sample size $n$, tuning parameter $\gamma$, and signal strength. Following the approach in \citet{zhang2020bootstrap}, we generate data from the logistic regression model: 
\begin{equation}
\label{model-logit}
\pi_i = \Pr(y_i=1|\mathbf{x}_i)= \mbox{expit}(\beta_0+\beta_1 x_{i1}+...+\beta_5 x_{i5}),
\end{equation}
where the predictors $x_{ij}$ are independently simulated from a uniform $[0, 1]$ distribution. We consider four settings by varying the signal strength: $\bm{\beta}= \left( \beta_j \right) = (1, 2, -2, 2, -2, -2)^T$ versus $\bm{\beta}=(0.3, -0.5, 0.5, -0.5, 0.5, -0.5)^T$, and the sample size: $n=1,000$ versus $n=200$. For each setting, 200 datasets are generated. For each generated dataset, risk scores are obtained from the logistic model (\ref{model-logit}), and our method for conservative decisions is applied. Additionally, an independent test sample of size $n'=500$ is generated to evaluate the predictive power and calculate accuracy, coverage, and selective risks. 

\afterpage{
\begin{figure}[H]
     \centering
     \begin{subfigure}[b]{1.0\textwidth}
         \centering
         \caption{Stronger Signal with $n=1,000$}
         \includegraphics[width=.85\linewidth, height=0.2\textheight]{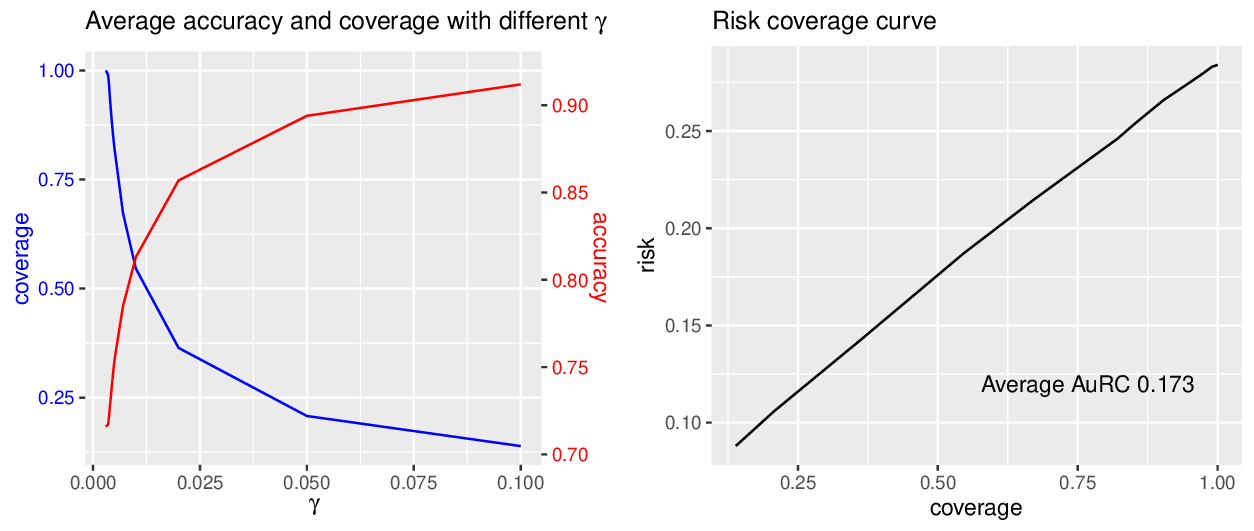}
         \label{fig3a}
     \end{subfigure}
     \begin{subfigure}[b]{1.0\textwidth}
         \centering
          \caption{Stronger Signal with $n=200$}
         \includegraphics[width=.85\linewidth, height=0.2\textheight]{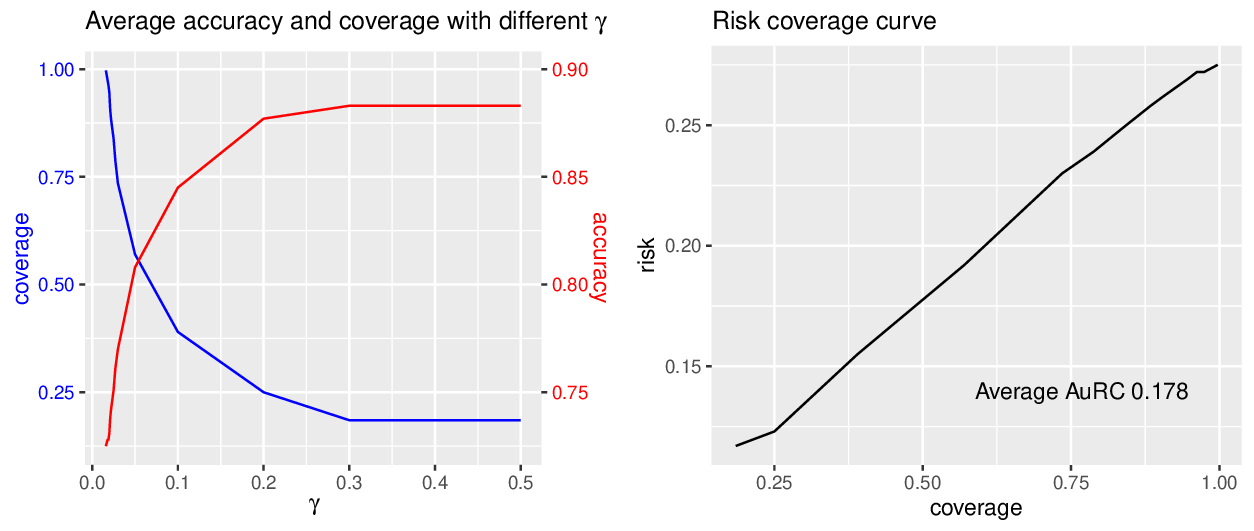}
         \label{fig3b}
     \end{subfigure}

     \begin{subfigure}[b]{1.0\textwidth}
         \centering
         \caption{Weaker Signal with $n=1,000$}
         \includegraphics[width=.85\linewidth, height=0.2\textheight]{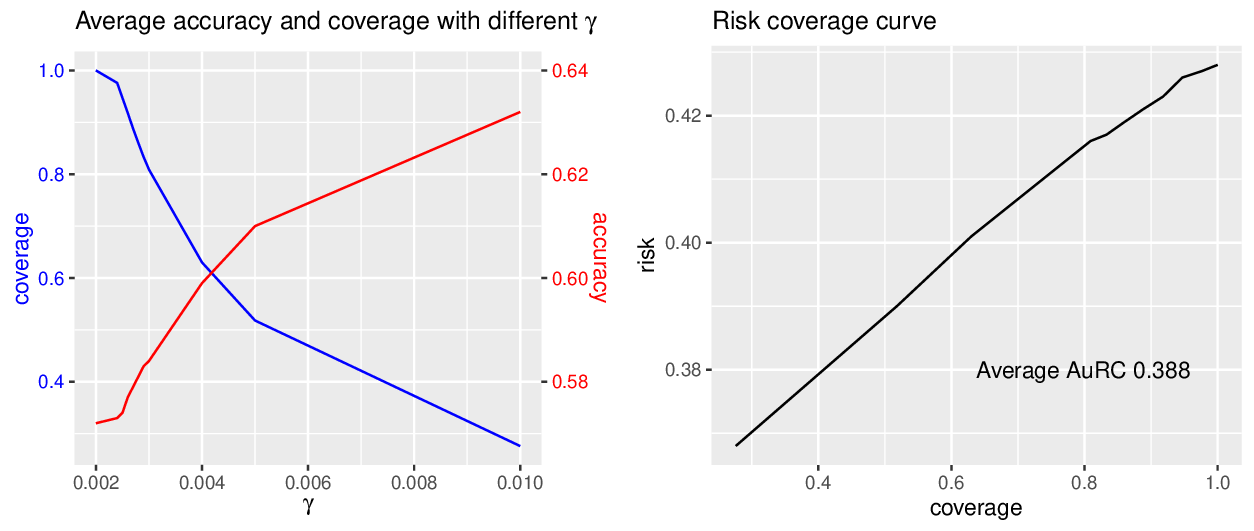}
        \label{fig3c}
     \end{subfigure}
     \begin{subfigure}[b]{1.0\textwidth}
         \centering
          \caption{Weaker Signal with $n=200$}
         \includegraphics[width=.85\linewidth, height=0.2\textheight]{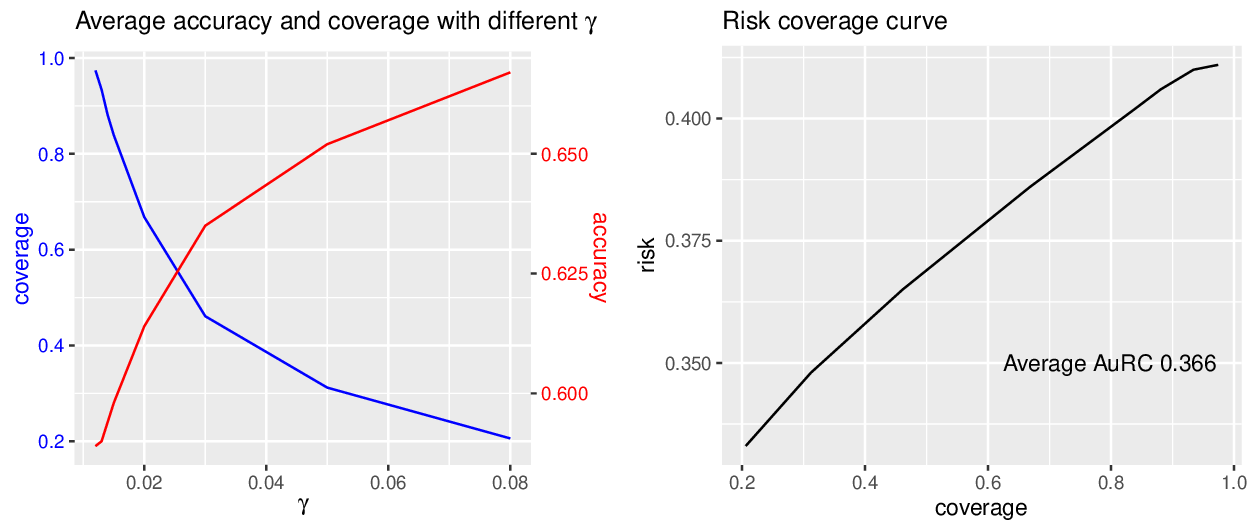}
       \label{fig3d}
     \end{subfigure}
        \caption{Results from Simulation Study 2 in Section \ref{sec-study2}. Each left panel plots the average risk, accuracy, and coverage plotted against varying $\gamma$ values. Each right panel provides the risk-coverage curve. \label{fig03}}
\end{figure}
\clearpage
}

Figure \ref{fig03} presents the averaged results from the 200 simulation runs. Two plots are made for each setting. The left panel plots the average accuracy and coverage for different values of the penalization factor $\gamma$. In the right panel, the averaged risk-coverage (RC) curve is displayed.  The RC curve plots the selective risk (misclassification error rate on accepted predictions) against the coverage. The average area under the RC curve (AuRC), as defined by \citet{franc2023optimal}, is computed to evaluate the model's overall performance. A smaller AuRC indicates better model performance. 

Increasing the penalty on classification error, represented by a larger value of $\gamma$, consistently leads to a decrease in coverage and an increase in accuracy across all scenarios. The simulation studies reveal several interesting observations. First, the effect of the penalty parameter $\gamma$ is influenced by the sample size. Smaller sample sizes necessitate a larger $\gamma$ to achieve a similar accuracy/coverage tradeoff, as shown in Figure \ref{fig03}. This is not surprising since the second part in the loss function Eq.~\ref{opt-softmargin} depends on the sample size. Secondly, Stronger signals exhibit a notable difference between the accuracy achieved on non-reject samples and the accuracy achieved on all samples. Conversely, weaker signals have a relatively smaller gap between these accuracies. This suggests that the rejection strategy is more effective with stronger signals. Thirdly, since the data are generated from a logistic model with a continuous event probability underlying the risk scores, the distribution of risk scores does not exhibit a gap, as seen in Study 1. Consequently, the plots do not show an `elbow' shape. In this situation, one may determine the optimal $\gamma$ by setting thresholds on coverage and accuracy. Fourthly, regarding optimization, there appears to be a sensitive region where coverage and accuracy exhibit drastic changes in response to variations in $\gamma$. Outside of this region, the changes are smoother. Consequently, fine-tuning of $\gamma$ is crucial within this sensitive region. 

\begin{figure}[H]
\centering
  \includegraphics[scale=.8, angle=0]{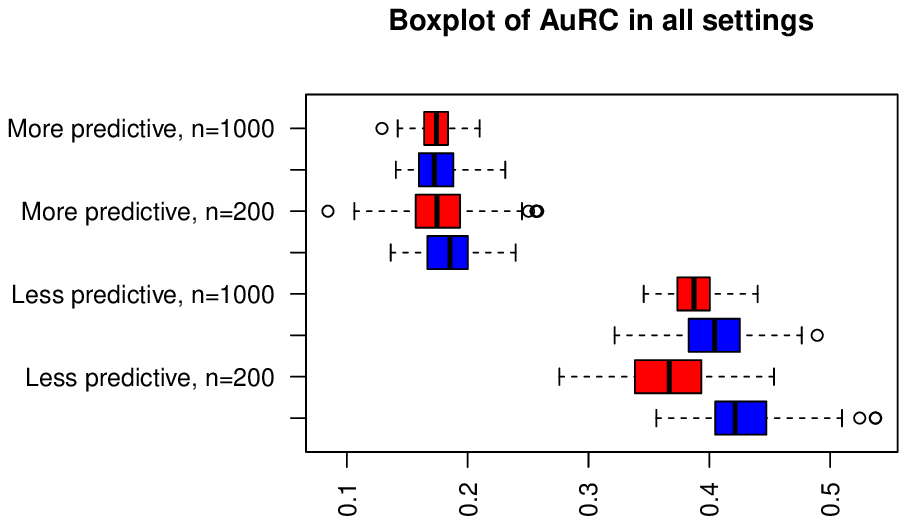}
  \caption{Boxplots of AuRC Values from Study 2 in Section \ref{sec-study2}. 
  For each setting, AuRC from training samples are plotted first, followed by AuRC values from test samples. \label{boxplot}}
\end{figure}

The area under the risk coverage curve (AuRC) has gained recognition as a recommended performance measure in recent literature. It provides a comprehensive assessment of the risk and coverage achieved by a classification model across different threshold values of $c$ and $d$. However, convergence issues may arise when the coverage approaches zero, leading to incomplete availability of the risk-coverage curve. In such cases, extrapolation methods are employed to calculate the AuRC. To further compare different settings, parallel boxplots of AuRC values are presented in Figure \ref{boxplot}. We can see a remarkable increase in AuRC as the signal weakens, which can be attributed to the vital influence of signal strength or the predictive power of the underlying risk score in determining AuRC. Additionally, the AuRC values on testing sets tend to be larger than those on training sets, as expected. Rejection models trained on smaller sample sizes exhibit greater variation in AuRC. These findings highlight the value of  AuRC as a reliable performance measure for classification problems.

To further illustrate the risk-coverage curve in conjunction with our method, we generate data from the same logistic model (\ref{model-logit}) with $\beta=(1, 3, -3, 3, -3, -3)$, where a total of $p=30$ predictors are all simulated from the $\mbox{Uniform}[0, 1]$ distribution. It is noteworthy that only the first five predictors actively contribute, while the others serve as noise variables.

\begin{figure}[H]
\hspace*{-2cm} 
\centering
\begin{subfigure}[b]{1.0\textwidth}
\centering
        \caption{}
         \hspace*{0.8in}
        \includegraphics[width=0.9\linewidth, height=0.28\textheight]{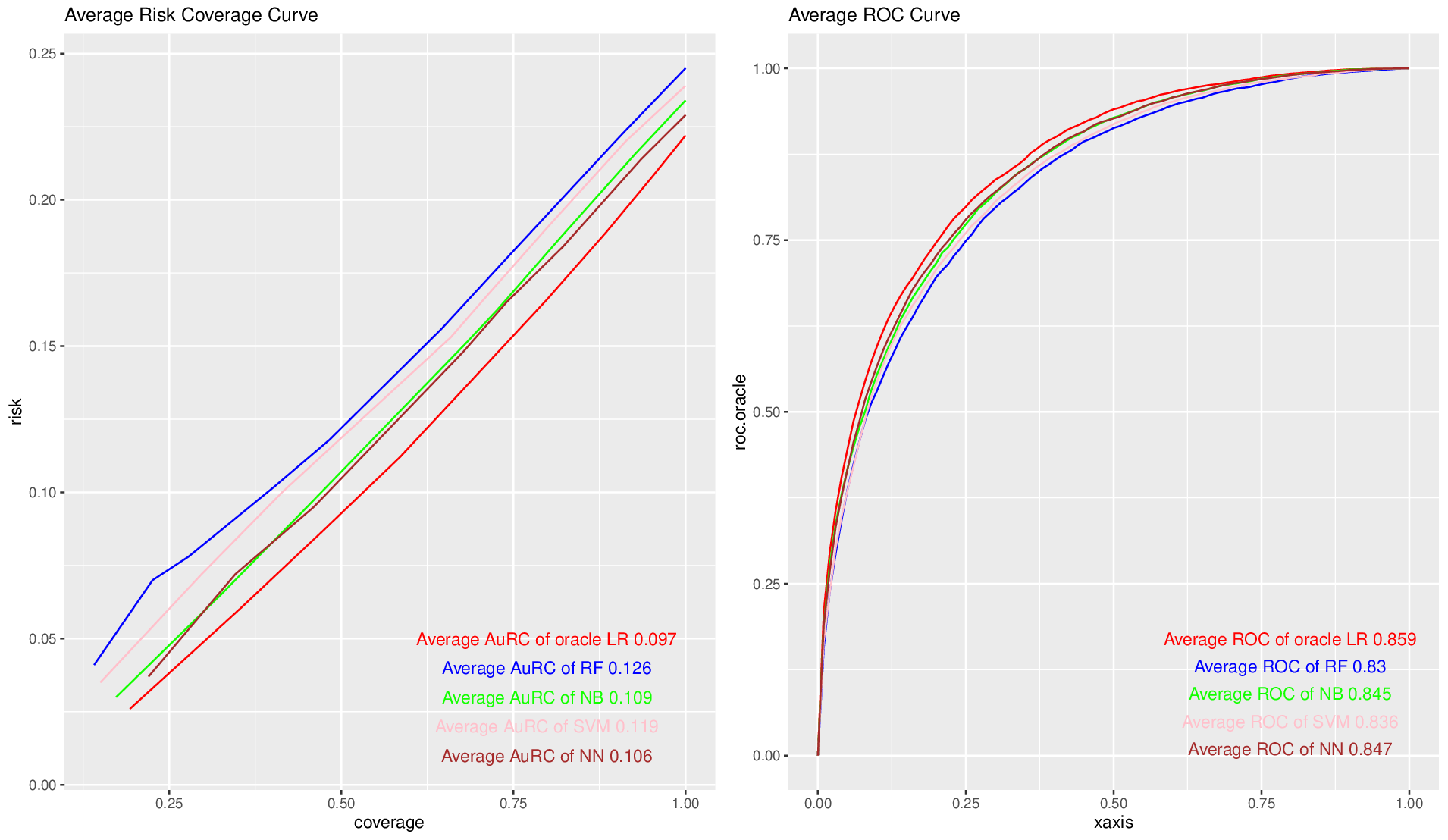}
         \label{fig5a}
     \end{subfigure}
        \begin{subfigure}[b]{1.0\textwidth}
         \centering
         \caption{}
      \includegraphics[width=1\linewidth, height=0.3\textheight]{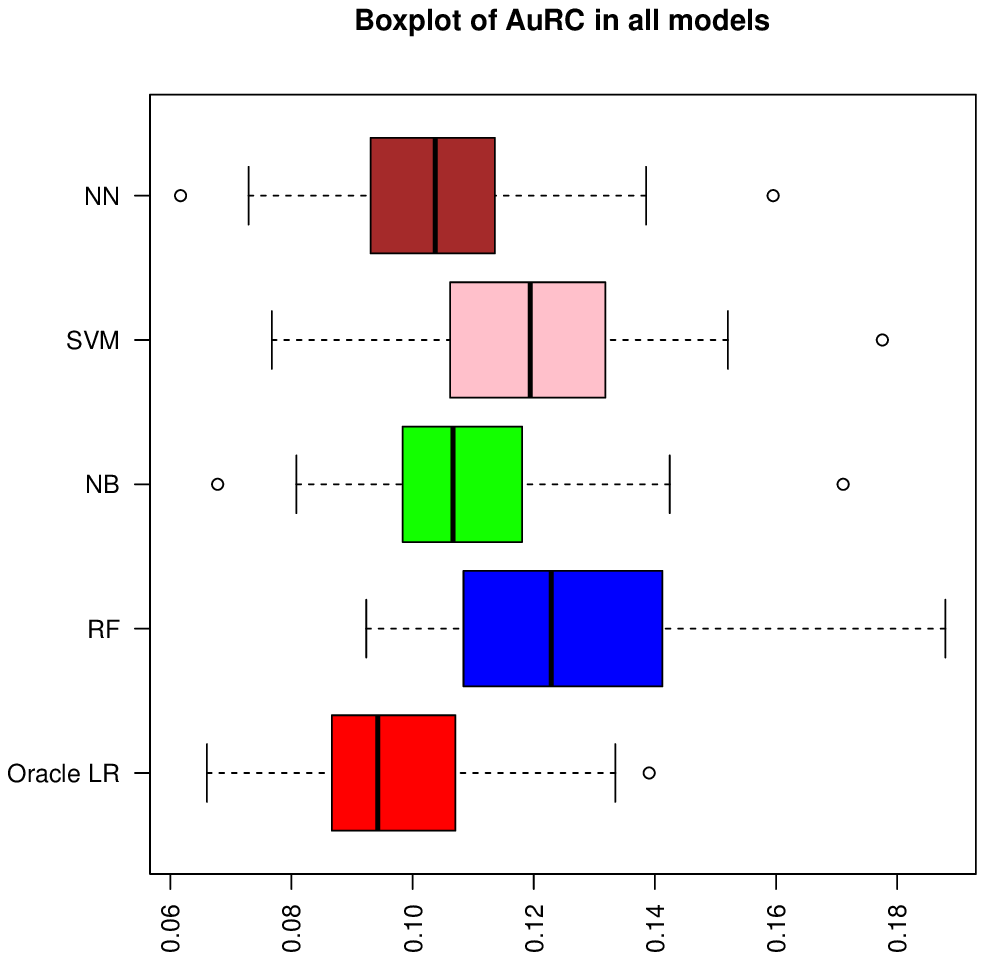}
         \label{fig5b}
     \end{subfigure}  
  \caption{Comparing classifiers through the risk-coverage (RC) curve and the ROC curve: (a) Averaged risk-coverage curve and ROC curve over 200 simulation runs; (b) Parallel boxplots of the area under the RC and ROC curves.  
  \label{fig5-classifers}}
\end{figure}

We experiment with only one sample size of $n=1,500$, where 1,000 observations constitute the training data, and the remaining 500 observations form the test data. For each simulated dataset, we train several classifiers \citep[see, e.g.,][]{hastie01statisticallearning}, including the oracle logistic model,  naive Bayes, SVM with radial basis function kernel, multilayer perceptron (MLP) neural network (NN), and random forest (RF)  using the training data. We then obtain predicted risk scores for the test data. Subsequently, we apply the optimal cutoff interval approach to compute the risk-coverage curve and the AUC measure. The oracle logistic model is fitted with only the first five predictors, while other classifiers are trained with all the covariates included, thus expecting potentially inferior performance. For each classifier, the default setting in its implementation package is used with minor tuning. 

Figure \ref{fig5-classifers} plots the summarized results over 200 simulation runs. Panel \ref{fig5a} displays the averaged RC and ROC curves while Panel \ref{fig5b} presents  the boxplots of AUC under the RC and ROC curves. 
It is evident that all classifiers exhibit strong performance in this classification task. As anticipated, the oracle logistic regression model demonstrates the highest performance, followed by NN, naive Bayes, and SVM. Random forest, on the other hand, performs less effectively, partly because of its limitation in handling linearity. 

In Panel \ref{fig5a}, both RC and ROC curves offer a comprehensive comparison of the classifiers and highlight the superiority of the oracle logistic model (depicted in red). However, the RC curve appears slightly advantageous, revealing clearer distinctions compared to the ROC curve. This conclusion is further supported by Panel \ref{fig5b}, where the oracle logistic regression model exhibits lower AUC values under the RC curves and higher AUC values under the ROC curves than the other classifiers.

\subsection{Study III: Two-step SVM}
\label{sec-study3}

This simulation study  serves as an illustrative example where re-fitting models on non-rejected observations can be beneficial. While the example specifically focuses on SVM, the conclusion can be extended to other methods as well. As Corollary \ref{corollary} suggests, in most cases, such iterations may not be required. However, when the data points are affected by a noise distribution, iterating between risk score estimation and cutoff interval can effectively eliminate the contaminating distribution and improve the recovery of the true decision boundary. 

\begin{figure}[H]
\centering
  \includegraphics[scale=1.0, angle=0]{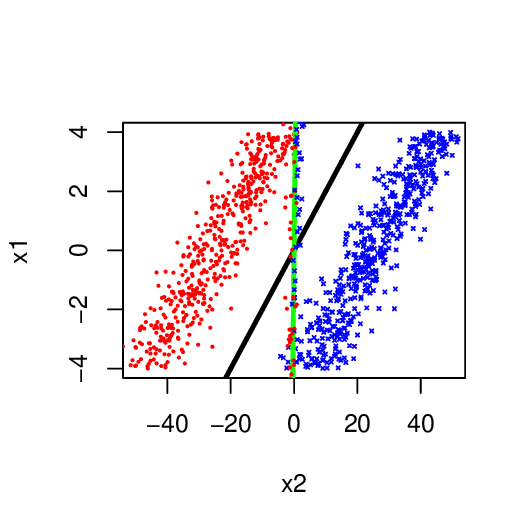}
  \caption{Simulated Data for Study 3 in Section \ref{sec-study3}. \label{setting3.dat}}
\end{figure}

Figure \ref{setting3.dat} illustrates the generation of data. The data set consists of two sets of observations with two covariates, $x_1$ and $x_2$, where $x_1 \sim \mbox{Uniform}(-4, 4)$. The first set, with a size of $n_1=1,000$, is simulated along the line $w_1 x_1 + w_2 x_2 + b =0$ 
with $(w_1, w_2, b) = (-5, 1, 0)$, where the random distance to the center line follows a normal distribution $N(5, 1)$. Observations above the line are positive, while observations below the line are negative.The second set, consisting of $n_2=200$ observations, represents noise and is simulated along the line $x_2=0.1x_1$, with a random distance to the central line following $N(1, 1)$.  

\begin{figure}[H]
     \centering
     \begin{subfigure}[b]{0.47\textwidth}
         \centering
         \caption{}
         \includegraphics[scale=.7,width=\textwidth]{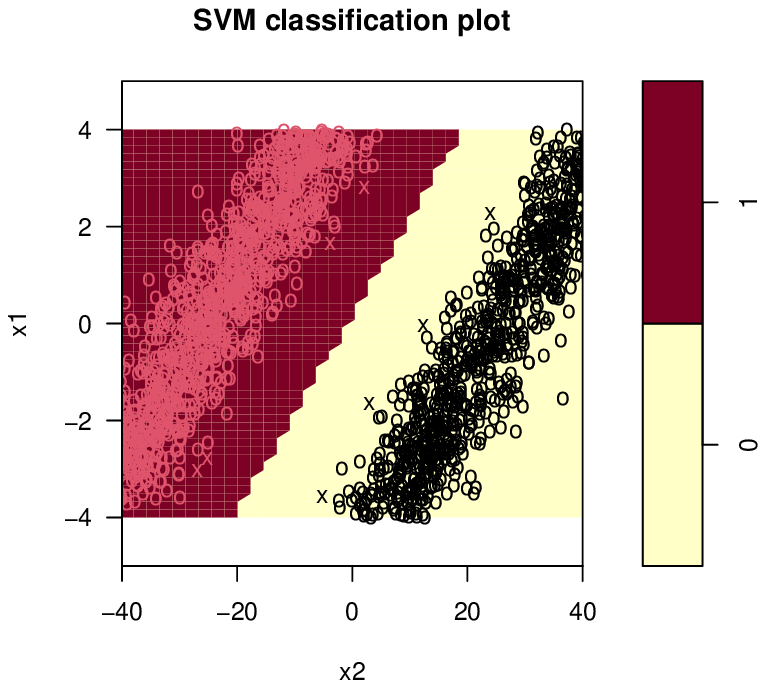}
         \label{fig:fit_on_dat1}
     \end{subfigure}
     \hfill
     \begin{subfigure}[b]{0.47\textwidth}
         \centering
         \caption{}
         \includegraphics[scale=.7,width=\textwidth]{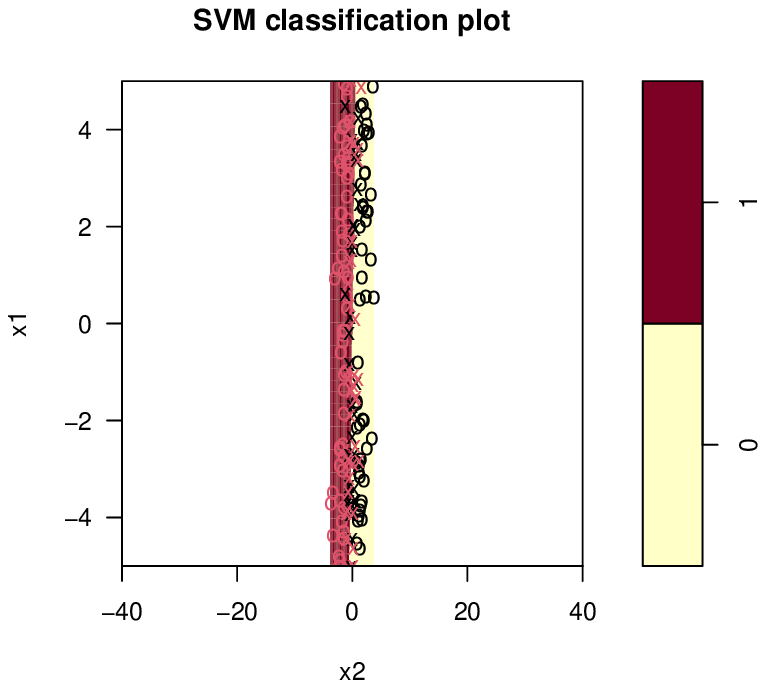}
         \label{fig:fit_on_dat2}
     \end{subfigure}

     \begin{subfigure}[b]{0.47\textwidth}
         \centering
          \caption{}
         \includegraphics[scale=.7,width=\textwidth]{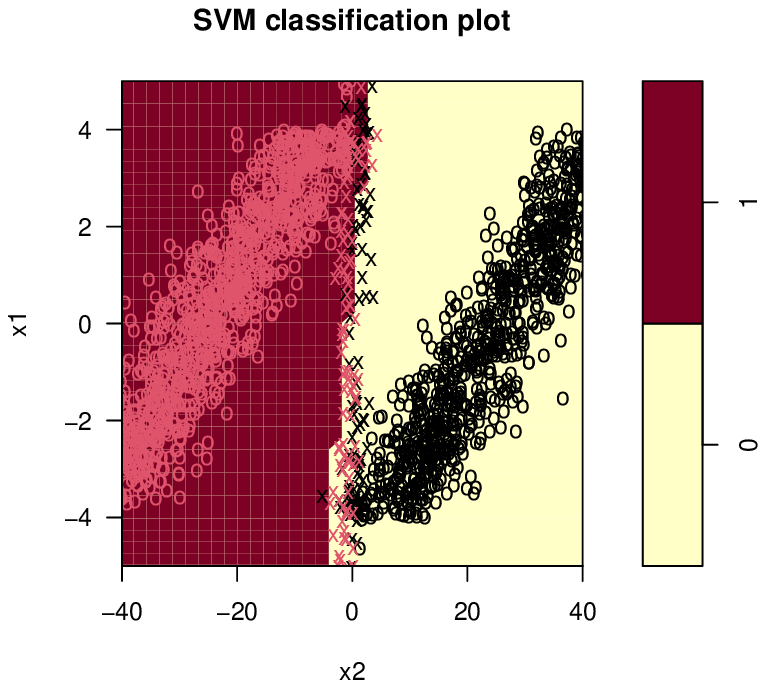}
         \label{fig:fit_on_all}
     \end{subfigure}
\hfill
     \begin{subfigure}[b]{0.47\textwidth}
         \centering
         \caption{}
         \includegraphics[scale=.7,width=\textwidth]{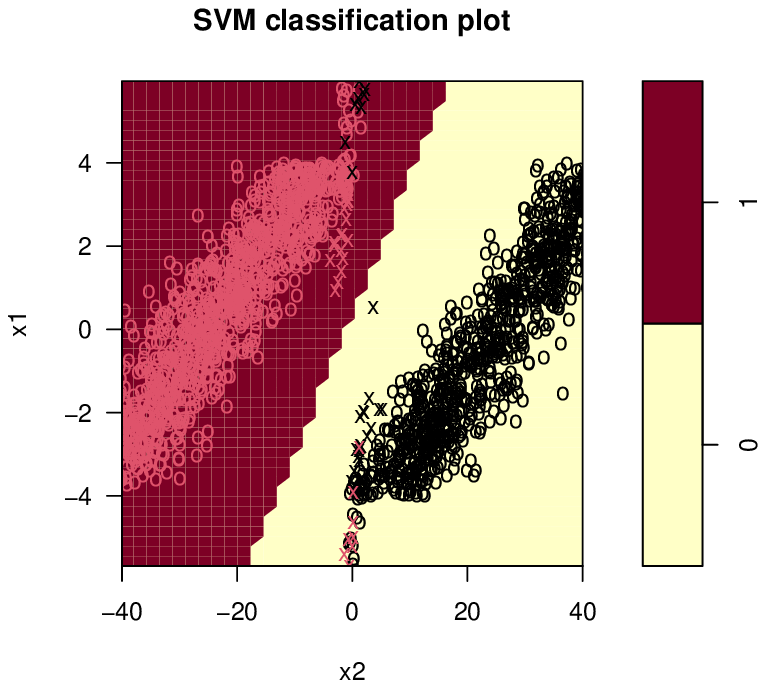}
         \label{fig:fit_on_reject}
     \end{subfigure}
        \caption{Two step SVM: (a) SVM fit on observations in group 1 only; (b) SVM fit on noise observations in group 2 only; (c) SVM fit on all data;  (d) SVM fit on non-rejected data.}
        \label{fig:two_step_svm}
\end{figure}

A two-step SVM approach is used, involving risk score estimation and cutoff interval estimation. In the first step, an SVM is trained on the complete dataset to estimate the cutoff interval $[c-d, c+d]$ using the fitted scores. In the second step, another SVM model is trained specifically on the non-rejected data points. Figure \ref{fig:two_step_svm} provides a visual representation of the two-step SVM approach using one data set. Panal (\ref{fig:fit_on_dat1}) displays a linear SVM fitted exclusively on the first set, representing the desired decision boundary. In Panal (\ref{fig:fit_on_dat2}, the SVM model is fitted on the second set of data, which consists of noise. Panal (\ref{fig:fit_on_all}) illustrates an SVM fitted on the entire dataset, where the noise distribution dominates the decision boundary despite comprising only 9.09\% (200/2,200) of the total data. Panal (\ref{fig:fit_on_reject}) depicts the decision boundary of the SVM fitted on non-rejected instances, obtained in the second step of the two-step SVM. The rejection band effectively eliminates most of the noise, resulting in a decision boundary that closely resembles the one in Panel (\ref{fig:fit_on_dat1}).

Table \ref{two_step_svm} summarizes the parameter estimates $\{\hat{w}_1, \hat{w}_2, \hat{b}\}$ obtained from 1,000 simulation runs. It can be seen that the regular SVM exhibits substantial bias in parameter estimation. In contrast, the two-step SVM approach effectively reduces this bias.

\begin{table}[H]
\caption{Results from Simulation Study 3 in Section \ref{sec-study3}. The parameters are normalized by $\parallel \bm{w} \parallel = \sqrt{w_1^2+w_2^2}$, and the estimates are similarly normalized.}
\vspace{.2in}
\centering
\scalebox{1.0}{
\begin{tabular}{lcccc} \hline \hline
 Method & Parameter & True Value &   Bias   &   Standard Deviation   \\ \hline
Regular SVM & $w_1$& 0.981 & $-0.826$ & 0.033 \\ 
 & $w_2$& $-0.196$ & $-0.791$ & 0.005 \\ 
 & $b$& 0 & $-0.006$ &0.107 \\ \hline
Two-Step SVM & $w_1$& 0.981 & $-0.014$ & 0.303\\ 
& $w_2$& $-0.196$ & $-0.049$ & 0.060\\ 
& $b$ &0& $-0.019$ & 0.919\\ \hline
SVM on Group 1 Only & $w_1$& 0.981 & $-0.006$ & 0.009\\ 
& $w_2$& $-0.196$ & $-0.025$ & 0.034 \\ 
& $b$ &0& 0.023 & 0.330\\ \hline
\end{tabular}
}
\label{two_step_svm}
\end{table}

\section{Application in Prostate Cancer Diagnosis}
\label{sec-example}

In a recent study \citep{gao2019application}, a new risk score was proposed based on the analysis of urinary volatile organic compounds (VOCs). This research was inspired by the fact that trained dogs are able to accurately detect prostate cancer by `sniffing' urine samples from patients, and the odor perceived by the dogs is produced by VOCs.  The new risk score was developed from a regularized logistic model that regresses the PCa status on VOC variables only; 
see \citet{gao2019application} for details. This new risk score demonstrated high potential as a reliable method for diagnosing prostate cancer by showing high sensitivity and specificity. 

\begin{figure}[H]
     \centering
 \begin{subfigure}[b]{0.47\textwidth}
         \centering
         \caption{}
         \includegraphics[scale=.7,width=\textwidth]{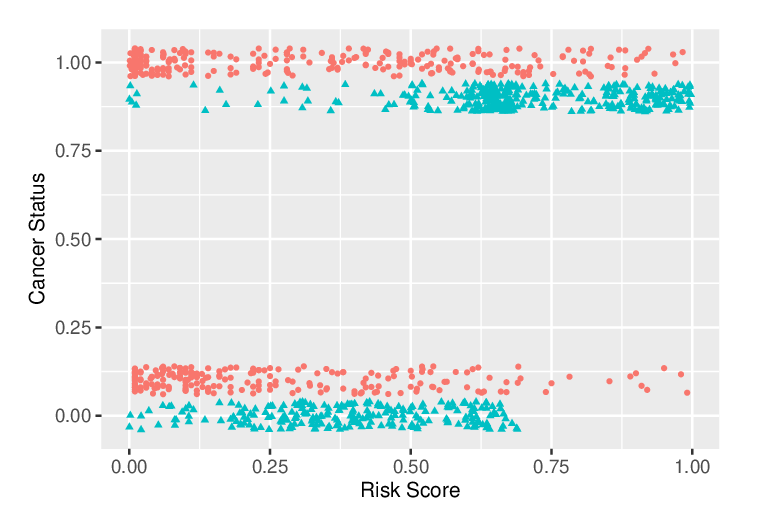}
     \end{subfigure}
     \hfill
     \begin{subfigure}[b]{0.47\textwidth}
         \centering
         \caption{}
         \includegraphics[scale=.7,width=\textwidth]{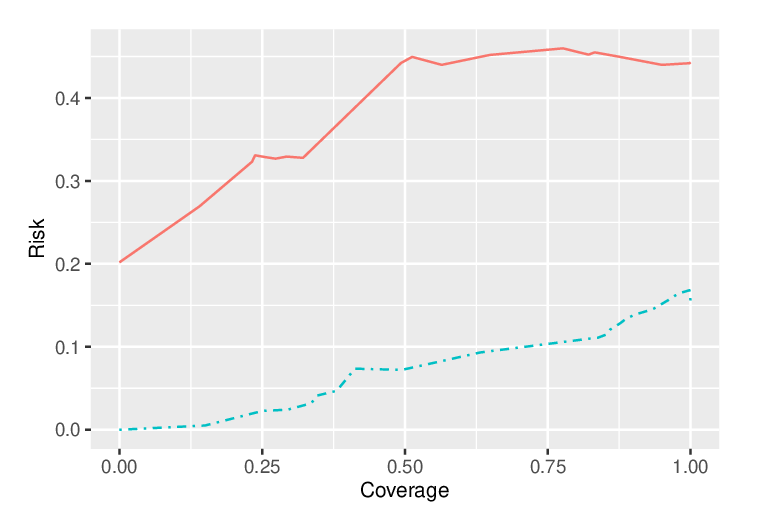}
     \end{subfigure}

     \begin{subfigure}[b]{0.47\textwidth}
         \centering
          \caption{}
         \includegraphics[scale=.7,width=\textwidth]{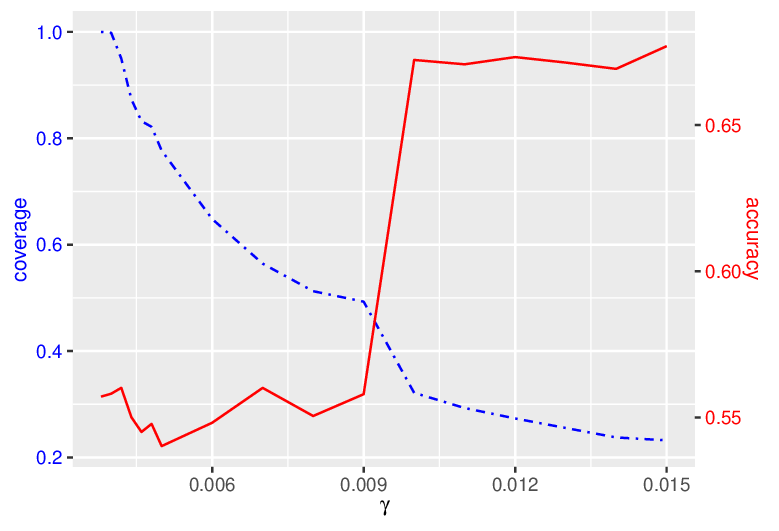}
     \end{subfigure}
\hfill
     \begin{subfigure}[b]{0.47\textwidth}
         \centering
         \caption{}
         \includegraphics[scale=.7,width=\textwidth]{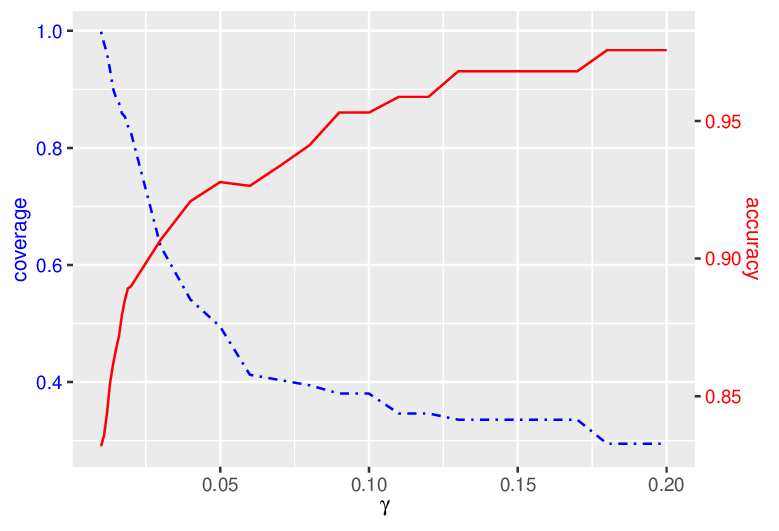}
     \end{subfigure}
        \caption{Prostate cancer diagnosis based on PSA and VOC-based risk scores. Panel (a) presents a jittered scatterplot of risk scores (PSA/10 in dots and VOC in triangle) against PCa status, PSA greater than 10 are excluded for visual performance; Panel (b) displays the risk-coverage curves, with AuRC values of 0.40 for PSA (solid line) and 0.07 for VOC-based risk scores (dashed line); Panels (c) and (d) depict coverage (dashed line) and accuracy (solid line) variations versus $\gamma$ for PSA and VOC-based risk scores, respectively. \label{fig07-PCa}}
\end{figure}

The data used in our analysis is an expanded version of the study conducted by \citet{gao2019application}, with a larger sample size. The dataset consists of 560 subjects, out of which 328 (58.6\%) have been confirmed as positive for PCa (prostate cancer). It is important to note that the study population comprises adult males who either displayed symptoms resembling those of PCa or were suspected to have the condition. Consequently, the PCa positive rate within this population is much higher than the prevalence rate observed in the general public. For each subject, both PSA and VOC-based risk scores, along with  the corresponding cancer status, have been recorded. Figure \ref{fig07-PCa}(a) presents a jittered scatterplot of the PSA and VOC-based risk scores versus the cancer status. For better visual performance, PSA greater than 10 is truncated. The figure clearly demonstrates that the VOC risk score exhibits a much stronger discriminative power compared to PSA.

We evaluated both the prostate-specific antigen (PSA) and the new VOC-based risk score using our proposed method. The risk-coverage plots are shown in Figure \ref{fig07-PCa}(b). Notably, the VOC-based risk score outperforms PSA with remarkably lower misclassification error rates. The area under the risk-coverage curve (AuRC) values for PSA and the VOC-based risk score are 0.40 and 0.07, respectively. In conclusion, this novel approach could potentially offer a more accurate and noninvasive diagnostic tool for prostate cancer diagnosis.

In Figure \ref{fig07-PCa}, Panel (c) plots both coverage and accuracy versus the tuning parameter $\gamma$ for PSA, and Panel (d) presents the same plot for the VOC-based risk scores. This plot serves as a valuable tool for selecting the optimal value of $\gamma$. For PSA, a widely adopted threshold of 4 or above yields a poor accuracy of 54\%.  With the proposed method, the baseline accuracy without any rejection is 55\%. Interestingly, there is a significant leap in accuracy around $\gamma = 0.01$. At this point, the accuracy can reach up to 68\%. However, it is important to note that this improvement in accuracy comes at the cost of reduced coverage, which is approximately 25\% only. For the VOC-based risk score, the plot exhibits a similar pattern to what has been observed in our simulation studies. Specifically, there is an `elbow' shape in the plot around $\gamma = 0.13$. By selecting this value of $\gamma$, an accuracy of 97\% is achieved, but with a relatively low coverage of 34\%. This level of coverage may not be desirable in practical applications. To ensure a coverage 80\% or above, one may select $\gamma = 0.02.$ With this choice of $\gamma$, we have an accuracy of 0.89. The resultant optimal cut interval is $(0.470, 0.616)$ with $c = 0.543$ and $d =0.073.$

The identified $c$ and $d$ is also compared with with with the bootstrap cutpoint CI proposed in \citet{zhang2020bootstrap}. In particular, the \texttt{method = MaxEfficiency} option is used as cutoff point selection criteria in \citet{zhang2020bootstrap} and the bootstrap size is set to be 200. IJ-based CI with correction method gives the confidence interval as [0.57, 0.73] (middle point $0.65$ and half interval length $0.08$). With this choice, the accuracy is 0.873 with the coverage of 0.663. With the same level of coverage, the proposed method achieves accuracy at 0.903. Comparatively, the cutpoint CI quantifies uncertainty in estimating the optimal cutpoint, while our proposed method directly minimizes risk while maintaining a specified coverage.

\section{Discussion}
\label{sec-discussion}

We proposed a novel approach to conservative decision-making for risk scores within the bounded coverage framework of \citet{franc2023optimal}. We formulated the optimal cutoff interval as a convex programming problem, leveraging similarities to SVM. However, our approach differs from SVM in the context of conservative decision-making where the objective is to minimize the classification margin instead of maximizing it. We have also obtained the theoretical solution under mild assumptions. The theoretical solution is essentially determined by the conditional probability of the response being positive, i.e., $\Pr(y=1|r).$ This aligns well with previous findings in the literature. As noted in \citet{herbei2006classification} and \citet{bartlett2008classification} , accurately estimating the conditional probability significantly affects the accuracy of the estimated rejection region in the cost-based framework. However, our proposed method circumvents the need for explicitly estimating the underlying probability. 

From a modeling perspective, it is natural to refine the model based on the non-rejected observations only. The extent of improvement achieved with the refitted model is contingent upon the underlying data distribution. In Simulation Study 3, we explored this idea and presented a scenario where  the rejection of uncertain instances led to a noteworthy enhancement in the accuracy of the fitted model. However, it should be noted that the performance improvement can be marginal in other cases, depending on the specific circumstances and data characteristics.

The selection of the penalty parameter, $\gamma$, is a significant consideration in practical applications. In our simulations, Study 1 demonstrates ideal scenarios where a specific choice of $\gamma$ is available. The optimal value of $\gamma$ is typically associated with an `elbow' shape observed in the accuracy and coverage curve as $\gamma$ varies. However, it should be noted that the `elbow' shape may not be applicable in all scenarios, as previously mentioned. In such cases, the choice of $\gamma$ may be based on the desired coverage level. In practice, achieving a coverage greater than 80\% or 90\% is often of interest. Additionally, the risk-coverage curve, generated by varying $\gamma$ across all possible values, serves as a valuable performance measure for comparing different risk scores or classifiers. It provides informative insights into the trade-off between risk and coverage. In terms of optimization, our limited experiences suggest the existence of a sensitive region for $\gamma$. Within this region, even a small adjustment in $\gamma$ can result in a noteworthy impact on accuracy or coverage. However, outside of this region, the changes in accuracy and coverage may not be as pronounced when $\gamma$ varies. Consequently, it is necessary to fine-tune the penalty parameter, $\gamma$, within the sensitive region.  

Our proposed method focuses on one-dimensional risk scores that exhibit correlation with the underlying conditional probability of $y$ being positive. This approach is comparable to the `plug-in' method as discussed in \citet{herbei2006classification} and \citet{franc2023optimal}. One advantage of our method is its flexibility in that the risk score can be obtained from various classification tools. This ensures its wide applicability across diverse scenarios. Additionally, within the framework of bounded coverage or bounded improvement, many modeling tools can be similarly modified. For instance, SVM can be reformulated to classify $y$ based on multi-dimensional $\bm{x}$ while maintaining a bounded coverage. Exploring a comprehensive approach that can simultaneously estimate the risk score and rejection region would be a promising direction for future research efforts. Along the similar lines, extending the method to encompass multiclass classification could be explored in future work.

\if1\blind
{
\section*{Acknowledgements}

This research was partially supported by National Cancer Institute of the National Institutes of Health (NCI-NIH) under Award Numbers: 1T32GM144919, R25GM069621, SC1CA245675 and 2U54MD007592. We express our gratitude to Dr.~Sabur Badmos and Ms.~Elizabeth Noriega Landa for their valuable contributions in collecting and preparing the prostate cancer data.
} \fi

\bibliographystyle{apalike}
\bibliography{References}


\FloatBarrier

\clearpage
\newpage
\clearpage \setcounter{page}{1} \setcounter{table}{0}
\setcounter{figure}{0}
\renewcommand{\thetable}{\Roman{table}}
\renewcommand{\thefigure}{\Roman{figure}}
\numberwithin{equation}{section}

\begin{center}
{\Large\bf SUPPLEMENTARY MATERIALS  \\ \vspace{0.1in} 
Conservative Decisions with Risk Scores } \\
\vspace{0.25in}
\textbf{Yishu Wei}, \textbf{Wen-Yee Lee}, \textbf{George Ekow Quaye}, and \textbf{Xiaogang Su}
\end{center}

\section{Proofs}
\noindent \textit{Proof of Theorem 1}:~~  
The risk function is
\begin{eqnarray*}
 R(\widehat{y}_{c,d}) &=& E_{y, r} l(y, \widehat{y}_{c,d}) \\
 &=&  E_{y, r} \left[ 0 \cdot I(|r-c| \leq d) + I \left( |r-c| > d \right) \cdot I \left\{ y (r-c) \leq - d  \right\}  \right] \\
 &=& E_r E_{y|r} \left[ I \left( |r-c| > d \right) - I \left( |r-c| > d \right)\cdot I \left\{ y (r-c) > - d \right\} \right] \\
 &=& \Pr( |r-c| > d) - E_r  \left[ I \left( |r-c| > d \right) \cdot E_{y|r} I \left\{ y (r-c) > - d \right\} \right] \\
 &=& \theta - E_r \left[ I \left( |r-c| > d \right) \cdot E_{y|r} I \left\{ y (r-c) > - d \right\}  \right],
\end{eqnarray*}
where the term $E_{y|r} I \left\{ y (r-c) > - d \right\}$ equals
\begin{eqnarray}
&=& E_{y|r} \left[ I(y=+1) I\left( r > c+d \right) + I(y=-1)  
I \left( r < c - d \right)  \right] \nonumber \\
&=&   \pi(r) I(r > c+d ) + (1- \pi(r)) I(r < c-d) \label{conditional-accuracy}
\end{eqnarray}
with $\pi(r) = \Pr(y=+1|r).$ Thus we have 
\begin{eqnarray*}
 R(\widehat{y}_{c,d}) & =& \theta - E_r \left[ I \left( |r-c| > d \right) \cdot \left\{  \pi(r) I(r > c+d ) + (1- \pi(r)) I(r < c-d)  \right\}  \right] \\
 & =& \theta - E_r \left[ I \left( |r-c| > d \right) \cdot \left\{  \pi(r) I(r > c ) + (1- \pi(r)) I(r < c)  \right\}  \right]
\end{eqnarray*}
From the above form, we would like to point out that the arguments of \citep{franc2023optimal} are not applicable here. Although it is known that 
the Bayes classifier $$ \widehat{y}_B= I(\pi(r) \geq 0) = I(r \geq c^\star) = \widehat{y}_{c^\star, 0}$$ maximizes the term $ \pi(r) I(r > c ) + (1- \pi(r)) I(r < c)$ among all classifiers $\widehat{y}_{c,d}$. The coverage function $I \left( |r-c| > d \right) $ varies with $c$ and $d$ for different classifiers $\widehat{y}_{c,d}.$ 

To proceed, we rewrite $R(\widehat{y}_{c,d})$ using (\ref{conditional-accuracy}),  
\begin{eqnarray*}
 R(\widehat{y}_{c,d}) &=& \theta - E_r \left[ E_{y|r} I \left\{ y (r-c) > - d \right\}  \right] \\
 &=& \theta - E_r \left[  \pi(r) I(r > c+d ) + (1- \pi(r)) I(r < c-d) \right]. 
\end{eqnarray*}
It suffices to show that $(c^\star, d^\star)$ maximizes  $E_r h(r, c, d)$ with $$ h(r, c, d) = \left[\pi(r) I(r > c+d ) + (1- \pi(r)) I(r < c-d)\right],$$ subject to $\Pr( |r -c| > d) = \theta.$ We have different scenarios to consider. 

First, consider the scenario when $ c^\star - d^\star < c^\star+d^\star < c-d < c+d$. Over the interval $( c^\star+d^\star, c-d)$, we have $r > c^\star$ and hence $\pi(r) > 1/2$. It follows that $h(r, c, d) = 1-\pi(r) < 1/2 < \pi(r) = h(r, c^\star, d^\star).$ 
 
Secondly, consider the scenario when $c^\star - d^\star < c^\star < c-d < c^\star+d^\star < c < c+d$. Then 
\begin{eqnarray*}
h(r, c^\star, d^\star) &-& h(r, c, d) ~=~ \int_{c^\star+d^\star}^{c+d} \pi(r) dF_r(r) - \int_{c^\star-d^\star}^{c-d} (1-\pi(r)) dF_r(r) \\
&=& \int_{c^\star+d^\star}^{c+d} \pi(r) dF_r(r) - \int_{c^\star-d^\star}^{c^\star} (1-\pi(r)) dF_r(r) - \int_{c^\star}^{c-d} (1-\pi(r)) dF_r(r) \\ 
&\geq & \int_{c^\star+d^\star}^{c+d} \pi(r) dF_r(r) - \int_{c^\star-d^\star}^{c^\star} (1-\pi(r)) dF_r(r) - \int_{c^\star}^{c-d} \pi(r) dF_r(r) \\ 
& =& \pi( r_1) \left[F_r(c+d) - F_r(c^\star - d^\star) \right] - 
\pi(r_2) \left[F_r(c^\star) - F_r(c^\star - d^\star) \right] \\
&& - \pi(r_3) \left[F_r(c-d) - F_r(c^\star) \right] \mbox{by mean value theorem} \\
& \geq & \pi( r_1) \left[F_r(c+d) - F_r(c^\star + d^\star) \right] - 
(\pi(r_2) \vee \pi(r_3)) \left[F_r(c-d) - F_r(c^\star - d^\star) \right] \\
&=& \left \{ \pi( r_1) - \pi(r_2) \vee \pi(r_3) \right\} \left[F_r(c+d) - F_r(c^\star + d^\star) \right] \\
& \geq & 0,
\end{eqnarray*}    
where the fact that $ 1- \pi(r) < \pi(r) $ when $r > c^\star$ is used in the third step; the mean value theorem for Lebesgue-Stieltjes integrals is applied in the fourth step with some $r_1 \in [c^\star + d^\star, c+d]$, 
$r_2 \in [c^\star, c^\star + d^\star]$ by symmetry of $\pi(r)$ around $c^\star$, and $r_3 \in [c^\star, c-d]$; in the sixth step, note that 
$$ F_r(c+d) - F_r(c^\star + d^\star) = F_r(c-d) - F_r(c^\star-d^\star) = \theta - \Pr(c-d < r < c^\star + d^\star);$$ 
and $\pi(r_1) \geq \pi(r_2) \vee \pi(r_3)$ by monotonicity of $\pi(r)$ in the last step. 

For other scenarios, similar approaches can be used to verify. The proof is completed. \qed
\vspace{0.2in}

\noindent \textit{Proof of Lemma 1}:~~~ Define $g(r) = \{\pi(r) - \pi(-r) \}/2.$ Then $g(r)$ is an odd function and hence satisfies the symmetry condition, yet at $r=0.$ Applying translation and scaling transformations to $g(r)$ yields $\pi'(r)$. By appropriately choosing the translation and scaling parameters, we can ensure that $\pi'(r)$ is symmetric around $c^\star$ and remains bounded within the interval $(0,1)$.

It remains to show the monotonicity. Given $\pi(r) \geq \pi(r'),$ it follows that $r \geq r'$ and $-r' \geq -r$ since $\pi(r)$ is monotone increasing. Consider 
$$ g(r) - g(r')  = \frac{\pi(r) - \pi(-r)}{2} - \frac{\pi(r') - \pi(-r')}{2} = \frac{1}{2} \left[ \{ \pi(r) - \pi(r') \} + \pi(-r') - \pi(-r) \right]  \geq 0.$$ 
Since the translation and scaling transformations applied to $g(r)$ are continuous and strictly increasing functions, they preserve the monotonicity of $g(r)$. Therefore, the function $\pi'(r)$ obtained from these transformations remains monotone increasing and satisfies the required symmetry and boundedness properties. This completes the proof. \qed

\section{Algorithm}
\label{algorithm}
We provide a brief describe of the algorithm. First, we solve the dual problem (Eq. 7) for the $\mu_i$ values  using, e.g., the R package \textbf{CVXR}  \citep{fu2017cvxr}. Then we follow steps outlined below to obtain $ l = c-d$ and $u = c+d$, which represent the lower and upper bound of the overlapping boundary respectively. Then the estimates of $c = (l+u)/2$ and $ (l-u)/2$ can be derived immediately. 

\IncMargin{1em}
\begin{algorithm}[H]
\caption{Cutoff Interval with Risk Score} \label{Algorithm-1}
\SetKwData{Left}{left}\SetKwData{This}{this}\SetKwData{Up}{up}
\SetKwInOut{Input}{input}\SetKwInOut{Output}{output}
\algrule 
\KwData{$\mathcal{D}=\left\{(r_i, y_i) \in \mathbb{R} \times \{\pm 1\} \right\}_{i=1}^n$ for binary classification} 
\KwResult{Center $c$ and half-width $d$} 
 Set the tuning parameter $\gamma >0$ \; 
\Begin{ 
        Set lower bound $l=\infty$ and upper bound $u=-\infty$\;
	Solve the dual problem (Eq. 7) for the $\mu_i$ values\;
	Search for the subset $S_1 = \{i: \mu_i=0\}$ of support vectors (SV)\;
	\hskip\algorithmicindent If $y_i=1$, then $l = r_i$\; 
	\hskip\algorithmicindent If $y_i=-1$, then $u=r_i$\;
	\If{both positive and negative SV are found in $S_1$} {Determine the values of $l$ and $u$ that yield the minimum $(u-l)$} 
	\Else {
	Search for the subset $S_2 = \{i: \mu_i=0\}$\;
	 \hskip\algorithmicindent $l \le r_i$ if $y_i=1$; 
     $u \ge r_i$ if $y_i=-1$\;
    Search for the subset $S_3$ of cases with $\mu_i=\gamma$\; 
	 \hskip\algorithmicindent $l \ge r_i$ if $y_i=1$; 
      $u \le r_i$ if $y_i=-1$\;    
	Determine the values of $l$ and $u$ that satisfy 
$$ \max_{i \in S_3, y_i = +1} r_i \leq l \leq \min_{i \in S_2, y_i = +1} r_i \mbox{~~and~~}  \max_{i \in S_2; y_i = -1} r_i \leq u \leq \min_{i \in S_3, y_i = -1} r_i $$ 
with minimum $(u-l)$\; 
}
Compute $c = (l+u)/2$ and $ (l-u)/2$\;
If no finite bounds can be found, conclude `non-convergence'.
} \algrule
\end{algorithm}

\vspace{.3in}
\noindent

\end{document}